# Utilising a Large Language Model to Annotate Subject Metadata: A Case Study in an Australian National Research Data Catalogue

**Shiwei Zhang[1], Mingfang Wu[2], Xiuzhen Zhang[1*]**

[1]RMIT University, Melbourne, Australia
[2]ARDC, Melbourne, Australia

## ABSTRACT

In support of open and reproducible research, there has been a rapidly increasing number of datasets made available for research. As the availability of datasets increases, it becomes more important to have quality metadata for discovering and reusing them. Yet, it is a common issue that datasets often lack quality metadata due to limited resources for data curation. Meanwhile, technologies such as artificial intelligence and large language models (LLMs) are progressing rapidly. Recently, systems based on these technologies, such as ChatGPT, have demonstrated promising capabilities for certain data curation tasks. This paper proposes to leverage LLMs for cost-effective annotation of subject metadata through the LLM-based in-context learning. Our method employs GPT-3.5 with prompts designed for annotating subject metadata, demonstrating promising performance in automatic metadata annotation. However, models based on in-context learning cannot acquire discipline-specific rules, resulting in lower performance in several categories. This limitation arises from the limited contextual information available for subject inference. To the best of our knowledge, we are introducing, for the first time, an in-context learning method that harnesses large language models for automated subject metadata annotation.

---

* Corresponding author: Xiuzhen Zhang (Email: xiuzhen.zhang@rmit.edu.au)

# 1. INTRODUCTION

In the last decade, there has been an increasing number of research data repositories where researchers and data providers can publish and share their research resources, such as datasets, data models, and software, to make them accessible and reusable for reproducible or new research. With the growing availability of datasets, it becomes increasingly challenging to discover the required datasets, whether from a repository or on the web. The discoverability of datasets largely depends on the quality of descriptive metadata, including information about a dataset's title, description, subject areas, authors, and geographical or temporal coverage. In this paper, we present a study on automated annotation of subject metadata to improve data discoverability. Subject metadata refers to terms from a pre-defined and well-organised vocabulary with terms describing or categorising subject areas of a resource; subject metadata facilitates data discovery by allowing users to browse subject areas and enhance the indexing of datasets, ultimately improving search quality. This approach provides multiple entry points for data discovery to support diverse user data search needs [1]. Currently, subject metadata of a dataset is primarily manually annotated by the dataset's owner, who are often expert researchers or data curators. As the volume of data continues to increase, the task of subject annotation becomes more labour-intensive. Cost-effective, automatic subject annotation using machine learning models has the potential to reduce the workload of manual annotation, as demonstrated in several recent studies [2, 3].

Machine learning models for adding subject metadata are primarily supervised classification models, which are driven by natural language processing (NLP) techniques. A classification model comprises two key elements: the features employed for training and inference, and the model responsible for capturing patterns inherent in these features. In the past decade, the field of NLP has undergone a significant shift in how machines comprehend text. This evolution has progressed from facilitating machines in grasping individual words or tokens through word embeddings e.g., word2vec [4], to more recently empower machines to comprehend sentences, paragraphs, or documents through the utilisation of contextual information, such as BERT [5]. BERT has introduced a new paradigm of learning machine models, which is to first pre-train a language model (LM) on large amounts of unstructured texts, and then fine-tune it on a downstream task, which has been becoming the dominant method in the NLP community. The emergence of the Generative Pre-trained Transformer (GPT) series of large language models (LLMs), notably GPT-3 [6], has ushered in a distinct machine learning paradigm with three sequential stages: LM pre-training, creation of a prompt, and provision of the tailored prompt alongside test examples to the LM to generate predictions, which is also referred to as in-context learning.

The primary advantage of in-context learning lies in its capacity to directly employ an LLM for a range of tasks, eliminating the need for training models. This is achieved by supplying the model with a tailored textual task instruction (zero-shot learning) and/or a small number of demonstration examples (few-shot learning), which is often referred to as prompt construction. Consequently, this alleviates machine learning practitioners from the resource-intensive burden of training machine learning models on costly hardware. With the progression of large language models (LLMs), often comprising billions of parameters, in-context learning has gained increasing attention within the NLP community due to its increasing effectiveness. Recently, GPT-4 has demonstrated performance comparable to that of humans across a range of professional and academic benchmarks [7]. As a result, there are a few recent studies that have investigated whether LLMs can be employed to replace humans as data annotators [8, 9, 10].

To the best of our knowledge, the potential application of LLMs for annotating subject metadata remains unexplored. In this paper, we fill this gap and propose to employ LLMs to annotate subject metadata for datasets from a research data catalogue. Building upon prior research [6, 11], we tailored prompts comprising the task instructions, demonstration examples, the classification rules, and the target record. We experimented with the GPT-3.5 which is the powerful LLM that drives the ground-breaking ChatGPT system† and the most cost-effective from the GPT-series of LLMs at the time when we conducted this research. We found that:
- GPT 3.5 demonstrates exceptional performance, achieving high precision in five subject disciplines. However, it also exhibits low performance in five disciplines.
- Additionally, including relevant demonstration examples in the prompt improves overall performance, with one high-precision discipline added and one low-precision discipline removed, but it does not improve performance across all disciplines.
- Due to the limited information provided in the context, in-context learning may constrain the model's ability to grasp the discipline-specific rules researchers from different disciplines apply when they submit datasets to data catalogues.

---

† https://chat.openai.com/

## 2. RELATED WORK

As our goal is to employ LLMs for automating the process of subject metadata classification, three research fields related to our work are automated subject metadata annotation, LLMs as annotators, and in-context learning.

### 2.1 Automated Subject Metadata Classification

Categorising academic journals, publications, and research-related resources like datasets into specific research disciplines constitutes a fundamental aspect of numerous bibliometric studies. Furthermore, this process contributes to enhancing the retrieval of information and expanding the accessibility of knowledge. The traditional method of annotating subject metadata is carried out manually by humans, whereas automated metadata annotation becomes feasible only when a large number of training data or labelling rules are accessible [12]. Various studies have explored the approach of training machine learning models on existing annotated metadata datasets and subsequently employ the trained models as automated annotators. For example, subject classification approaches for journals and papers frequently make use of clustering solutions that employ citation-based mapping approaches [13] and/or text-based mapping approaches [14]. Suominen et al. [3] introduced Annif, an open-source toolkit for automated subject indexing and classification, along with Finto AI, a service that utilises Annif for automated subject indexing. They make use of a variety of machine learning models, including Maui, fastText, Omikuji, and neural networks. Wu et al. [2] have investigated various classical machine learning approaches, such as KNN, SVM, MNB, and MLR, for automatically tagging subject metadata; their study discovered that automated methods perform effectively for well-represented subject categories, which generally possess distinctive features distinguishing them from other categories. All these conventional machine learning methods require a large set of training data that is usually lacking in research communities.

### 2.2 Using LLMs as Annotators

As the LLMs are becoming more capable and intelligent, they have the potential to be used as automated evaluators in machine learning tasks [15, 16]. Furthermore, recent studies have explored whether LLMs can serve as substitutes for crowdsourcing workers or contribute to lowering the expenses associated with dataset annotation. For example, Wang et al. [8] examined the application of GPT-3 as a data annotator, demonstrating that incorporating GPT-3 for annotating training datasets allows models to achieve similar performance levels of using human-annotated datasets across various natural language understanding and natural language generation tasks. Additionally, they observed that the cost of procuring labels from GPT-3 is substantially lower, ranging from 50% to 96%, compared to the expense of obtaining labels from human annotators. Gilardi et al. [9] have found that ChatGPT outperforms crowd-workers in several annotation tasks related to tweets. They also found that using ChatGPT incurs a cost twenty times cheaper than using Amazon Mechanical Turk. Instruction tuning is a new data-intensive training approach employed to substantially improve the zero-shot performance of LLMs on unseen tasks [17], or improve the truthfulness and reduce toxic output generation of LLMs [18]. Recently, Honovich et al. [19] utilised a variant of GPT-3 to generate a large number of instructions for fine-tuning other large language models. This approach exhibited superior performance compared to models trained on manually curated datasets across multiple benchmark evaluations. Interestingly, a recent study found that even 33–46% of crowdsourcing workers used LLMs when completing the annotation tasks [20]. While LLMs offer the possibility of reducing annotation expenses and time, they might not possess the capacity to completely substitute human annotators across all NLP tasks. This is due to the potential biases within the training data of LLMs and the challenges these models could face in accurately discerning intricate nuances and context-dependent meanings within language [10].

## 2.3 In-context Learning

GPT-3 [6] introduced a novel approach to machine learning, which dispenses the need for gradient updates or fine-tuning. Instead, task instructions and few-shot demonstrations are conveyed solely through textual interactions with the model. This method, known as in-context learning, enables GPT-3 to attain impressive results across various NLP tasks. In-context learning is derived from prompt-based learning, where the incorporation of prompt augmentation techniques is common with this approach [21]. As discussed in the introduction, the procedure of in-context learning comprises three steps, namely pre-training a LM, prompt engineering and inference. Pre-training an LM is the most expensive step, but currently, both public and private pre-trained LLMs are available. Machine learning practitioners now only need to customise prompts instead of going through the traditional machine learning procedure of training, validation, and deployment. For instance, prompt engineering has become the most impactful part that affects the performance of an in-context learning method. Early studies primarily revolve around manual customization of prompts, as seen in examples like [6], where tailored prompts are designed for different NLP tasks. While creating templates manually is intuitive and does enable the accomplishment of diverse tasks with a certain level of accuracy, this advancement requires both time and expertise. To address this issue, researchers have explored approximately two categories of automated prompt learning techniques: discovering hard prompts and tuning soft prompts. Hard prompts are human-readable and interpretable prompts that can be discovered by mining-based and paraphrasing-based methods [22]. Soft prompts or continuous prompts are geared towards tuning continuous vectors rather than individual words to serve as prompts. An example is prefix-tuning, which optimises a sequence of continuous task-specific vectors [23] that are added as a prefix to the input. In addition, demonstration examples play a significant role in in-context learning. Liu et al. [11] discovered that examples exhibiting semantic similarity to the test example yield superior results compared to randomly chosen instances. Incorporating chains of thought demonstrations as exemplars in prompts, as demonstrated by Wei et al. [24], has been observed to enhance performance across a spectrum of arithmetic, commonsense, and symbolic reasoning tasks.

## 3. METHODOLOGY

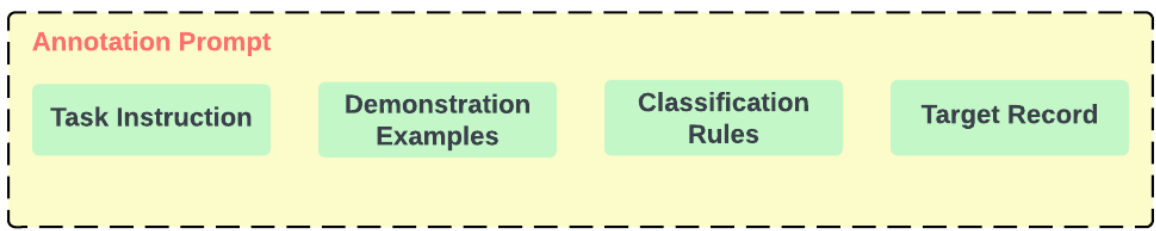

Figure 1: The overall structure of our prompt-based subject metadata annotation.

In the following section, we present our in-context learning method for the automatic classification of metadata records from Research Data Australia (RDA). RDA is a national research data cataloguing and discovery service offered by Australian Research Data Commons. Note that an RDA record contains various metadata. Using the title and description of each record, we design prompts to instruct LLMs to generate a subject label for the record. Our approach mainly utilises the extensive knowledge of LLMs to conduct inference. The primary challenge lies in designing a prompt that is specifically tailored to a specific task. Based on the prior research [6, 11], we design our prompts to have four components: the task instruction, demonstration examples, the classification rules, and the target record, as demonstrated in Figure 1.

## 3.1 Our prompts

> You are an assistant at Research Data Australia (RDA), and your task is to accurately determine the categories of a dataset given its title and description. Please categorize the given dataset into the divisions of Australian and New Zealand Standard Research Classification (ANZSRC):
>
> mathematical sciences / physical sciences / chemical sciences / earth sciences / environmental sciences / biological sciences / agricultural and veterinary sciences / information and computing sciences / engineering / technology / medical and health sciences / built environment and design / education / economics / commerce, management, tourism and services / studies in human society / psychology and cognitive sciences / law and legal studies / studies in creative arts and writing / language, communication and culture / history and archaeology / philosophy and religious studies
>
> **Examples of dataset classification**:
>
> 1. Dataset title: Mathematics of Cryptography   Dataset description: Mathematics of Cryptography. The Australian society and economy requires fast, reliable, and secure communication. First-generation security solutions are not capable of supporting the efficiency and scalability requirements of mass-market adoption of wireless and embedded consumer applications. New security infrastructures are emerging and must be carefully, but rapidly, defined. Thus developing new mathematically solid tools in this area is one of the most important and urgent tasks. Besides, the intended work advances our knowledge of the theory and the quality of our culture. As such, it will promote the Australian science and will also have many practical applications in Cryptography, Computer Security and E-Commerce.   Categories: mathematical sciences.
> .....
>
> **Note: Identify the relevant categories of the following dataset by examining its title and description. The answers should be limited to a maximum of three, separated by "/", and arranged in order of relevance, with the most relevant listed first.**
>
> **The following is information about the target dataset:**
> Dataset title: Towed video footage of the seafloor at Lorne, Victoria
> Dataset description: Observation data (towed video, BRUVs) collected in Victorian state waters at Lorne.This footage was collected by researchers from Deakin University, Victorian Department of Primary Industries - Marine and Freshwater Resources Institute (MAFRI) and Parks Victoria.The original footage has been converted from various formats including VHS and MiniDV to digital format, with funds supplied by Deakin University Library. Underwater footage gathered from other geographical locations around Victoria from the Victorian Marine Habitat Mapping Program can be accessed via the links featured at the bottom of this record.High quality versions of the videos may be requested via Deakin University Library.   Categories:

Figure 2: An example prompt for classifying an RDA record into ANZSRC divisions.

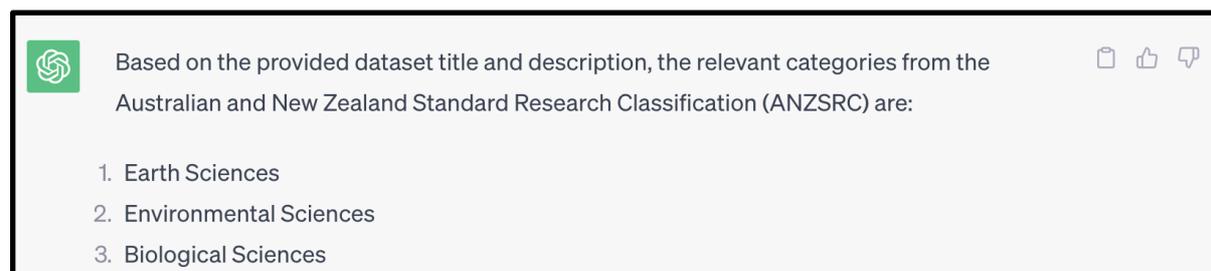

Figure 3: An example response from the interactive interface of GPT-3.5 or ChatGPT.

As shown in Figure 1, The task instruction illustrates the task that the LLM needs to perform and what the classification candidate labels are. The demonstration examples provide examples of what the input and output look like. The classification rules are designed for the LLM to follow during inference, with the aim of controlling its generation process. The final component is a testing record. The information on demonstration examples includes a title, the dataset's description, and an annotated label, while the testing record only has its title and description. We anticipate that the LLM will predict the potential category to which the target dataset may belong.

We use two different prompting strategies, and the primary distinction lies in the manner in which demonstration examples are chosen. The **GPT 3.5 - Random** prompt uses randomly selected examples for demonstration, while **GPT 3.5 - Relevant** prompt choses relevant examples for demonstration. In terms of selecting relevant examples, we use the embedding API of OpenAI[‡] to generate text embeddings for both target query and the existing dataset, and then utilising cosine similarity to find the most relevant (or highest cosine similarity scores) examples from the pool of demonstration examples. The demonstration examples of GPT 3.5 - Random primarily showcase the appearance of classifications and labels, while those of GPT 3.5 - Relevant consist of more relevant examples that encourage the model to engage in further in-context learning. Therefore, GPT 3.5 - Random tends to generate responses primarily based on its existing knowledge, whereas GPT 3.5 - Relevant incorporates more information from the given context along with its knowledge during the inference process. Both models have a set of 3 demonstration examples each. Figure 2 illustrates an example of our prompt, while Figure 3 presents the projected categories derived from the example in Figure 1 using the interactive interface of GPT-3.5. In our experiments, we use the API instead of the interactive interface of GPT-3.5 or ChatGPT, as the API provides more tools for controlling the response of GPT-3.5.

---

[‡] https://platform.openai.com/docs/api-reference/embeddings

In terms of selection of LLMs, we opt for GPT-3.5[§], which is considered the most cost-effective proprietary LLM as we addressed this problem. We utilise OpenAI's API[**] to send prompts and receive predictions. The following are the parameter settings used for the API. When it comes to selecting a specific GPT-3.5 model, we opt for "gpt-3.5-turbo" which offers the best cost-effectiveness and serves as the underlying engine powering ChatGPT. The combination of the temperature and max_tokens parameters offers a means to steer text generation. A higher value for temperature introduces more randomness into the generated text, while a higher value for max_tokens allows for longer sequences to be generated. To ensure the generation of concise and specific responses tailored to this task, we tuned these parameters and ultimately set the temperature to 0 and the max_tokens to 10.

```
response = openai.ChatCompletion.create(
    model="gpt-3.5-turbo",     # Specify the GPT-3 engine to use
    messages=messages,          # prompt
    temperature = 0,            # freedom of generation, 0 means greedy search
    n = 1,                      # number of returned responses
    max_tokens= 10              # Specify the maximum number of output tokens)
```

Here is an example of GPT 3.5's response to our query classification prompt, where the generated response or predictions from GPT 3.5 are stored in the "content" key's value.

```
{
  "choices": [
```

---

[§] https://platform.openai.com/docs/models

[**] https://platform.openai.com/

```json
    {
      "finish_reason": "stop",
      "index": 0,
      "message": {
        "content": "Environmental Sciences / Chemical Sciences.",
        "role": "assistant"
      }
    }
  ]
}
```

We perform post-processing on the generated responses, which includes tasks such as splitting the response, converting tokens to lowercase, and matching each token with the corresponding ANZSRC-FoR category. If the generated response does not include any ANZSRC-FoR categories, the default value "others" will be assigned.

## 4. DATA

Table 1: The statistics of RDA records

|  | #records |
|---|---|
| RDA dataset records | 209,294 |
| Dataset records with annotated ANZSRC-FoR | 86,233 |
| Dataset records without annotated ANZSRC-FoR | 123,061 |
| Testing Records | 1,684 (Maximum 100 record per category at the ANZSRC-FoR top two digits level) |

The RDA records are annotated with category labels from the Australian and New Zealand Research Classification—Fields of Research (ANZSRC-FoR)[††], which holds significant prevalence within the research community of Australia and New Zealand.

---

[††] https://www.abs.gov.au/statistics/classifications/australian-and-new-zealand-standard-research-classification-anzsrc/latest-release

This classification serves various purposes, including the reporting of research outputs. The ANZSRC-FoR has two versions: ANZSRC-FoR 2020 and ANZSRC-FoR 2008. Majority of RDA records are annotated using the ANZSRC-FoR 2008 version, which is structured hierarchically, comprising three levels distinguished by two-, four-, and six-digit codes. The top level, denoted by a 2-digit code, encompasses 22 divisions that encompass expansive research domains, e.g., "education", "economics", and "engineering." The second level, indicated by a 4-digit code, encompasses 157 groups, which further delineate specific research areas, e.g., "economic theory", "applied economics", and "econometrics." Finally, the third level, defined by a 6-digit code, encompasses a total of 1,238 distinct and specific research fields, e.g., "agricultural economics", "economic history", and "economics of education." In this study, we investigate the performance of in-context learning methods at classifying records at the top level or divisions with 2-digit code.

Table 1 presents the statistics of RDA records of dataset collections (RDA catalogues have other types of resources, e.g. projects. This study focused on the dataset records, which will be referenced as records in this paper). The number of current RDA records is 209,294, where 86,233 (~%41) of them have at least one ANZSRC-FoR category label. while the remaining 123,061 (~%59) doesn't have any subject metadata including ANZSRC-FoR annotations. This result in subject metadata is not used effectively in subject facet search and filter that are key features provided by the RDA discovery portal.

Table 2: The number of testing instances in each category. Fields are referred to as divisions in ANZSRC-FoR 2008, and code is short for ANZSRC-FoR 2008 2-digit code.

| Field | Mathematical sciences | Physical sciences | Chemical sciences | Earth sciences | Environmental sciences | Biological sciences | Agricultural and veterinary sciences | Information and computing sciences | Engineering | Technology | Medical and health sciences |
|---|---|---|---|---|---|---|---|---|---|---|---|
| Code | 01 | 02 | 03 | 04 | 05 | 06 | 07 | 08 | 09 | 10 | 11 |
| #records | 48 | 100 | 100 | 100 | 100 | 100 | 100 | 100 | 100 | 28 | 100 |

| Field | Built environment and design | Education | Economics | Commerce, management, tourism and services | Studies in human society | Psychology and cognitive sciences | Law and legal studies | Studies in creative arts and writing | Language, communication and culture | History and archaeology | Philosophy and religious studies |
|---|---|---|---|---|---|---|---|---|---|---|---|
| **Code** | 12 | 13 | 14 | 15 | 16 | 17 | 18 | 19 | 20 | 21 | 22 |
| **#records** | 100 | 85 | 44 | 34 | 100 | 36 | 7 | 100 | 100 | 100 | 2 |

A record may have one or more ANZSRC-FoR codes (for this study we focus on 2-digit codes, which encompass 22 research fields at high level). Over 90% of the records have a single code, while the remainder may have two or more. We focus on predicting the most relevant label for data records. If GPT-3.5's response contains multiple labels, we use only the first label in evaluation, which is the most relevant predicted label.

We randomly select records from each of the 22 fields to design and revise prompts, and we further separately sample records as test examples. We randomly chose 100 records from each category (note: some categories have less than 100 records) which ended up with about 1,455 RDA records as the pool for sampling demonstration examples used in the prompt. To construct testing examples, we randomly selected 100 records for each of the 22 ANZSRC divisions. As some fields, e.g., "technology," "law and legal studies," and "philosophy and religious studies," have fewer than 100 records, a total of 1,684 records are included in testing. The statistics of RDA records is presented in Table 2.

## 5. RESULTS and DISCUSSIONS

Table 3: The precision of two GPT 3.5 models and two best performing ML models from [2]. Categories containing fewer than 20 testing examples are excluded, as the reliability of the results for those categories may be compromised. Fields with exceptionally low performance (<=0.5) are highlighted in red, whereas fields with good performance (>=0.8) are highlighted in bold.

| Code | GPT 3.5 - Random | GPT 3.5 - Relevant | MLR | KNN |
| --- | --- | --- | --- | --- |
| 01 | 0.5 | 0.5 | 0.29 | 0.41 |
| 02 | 0.72 | **0.8** | **0.97** | **1** |
| 03 | **0.93** | **0.88** | 0.73 | 0.6 |
| 04 | 0.36 | 0.4 | **0.96** | **0.92** |
| 05 | 0.64 | 0.65 | 0.61 | 0.68 |
| 06 | 0.55 | 0.59 | 1 | 0.64 |
| 07 | **0.84** | 0.79 | 0.63 | 0.77 |
| 08 | 0.66 | 0.65 | 0.45 | 0.53 |
| 09 | 0 | 0.07 | **1** | **0.94** |
| 10 | 0 | 0.25 | 0.29 | 0.2 |
| 11 | 0.74 | 0.73 | 0.68 | 0.63 |
| 12 | **0.83** | **0.89** | 0.61 | 0.67 |
| 13 | **0.81** | 0.78 | 0.58 | 0.69 |
| 14 | 0.71 | 0.63 | 0.41 | 0.58 |
| 15 | 0.4 | 0.54 | 0.21 | 0.18 |
| 16 | 0.52 | 0.56 | 0.56 | 0.55 |
| 17 | 0.68 | 0.69 | 0.4 | 0.32 |
| 19 | 0.72 | **0.8** | **0.82** | 0.76 |
| 20 | 0.78 | **0.88** | **0.89** | 0.26 |
| 21 | 0.35 | 0.39 | **0.97** | **0.99** |
| Macro Average | 0.59 | 0.62 | 0.65 | 0.63 |
| Standard Deviation | 0.26 | 0.22 | 0.26 | 0.25 |
| Micro Average | 0.62 | 0.65 | 0.7 | 0.66 |

We employ precision as the metric to evaluate automated annotation. In a classification task, precision for a class is calculated by dividing the number of true positives by the total count of predicted positives. The value of precision reflects the quality of predictions. We can trust predictions for classes with high precision. Apart from reporting precision for each category, we also report both Micro average precision and Macro average precision. The results are presented in Table 3, where we also compare our methods with two supervised models from the previous study [2]. Since the precision of those two supervised models was achieved on a different testing dataset than ours, the results are not directly comparable. Nevertheless, we can still conduct an analysis to determine which categories the models find challenging.

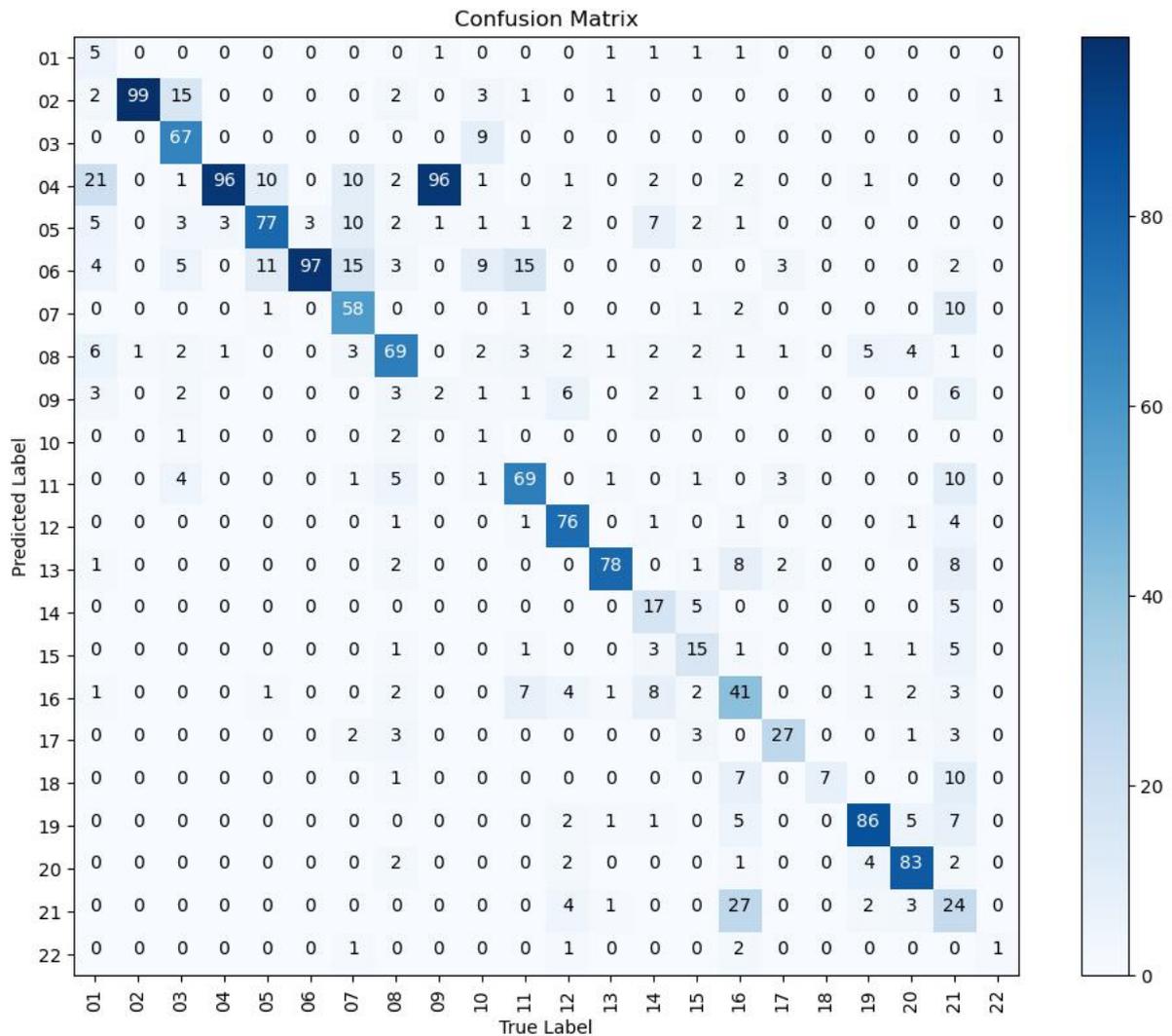

Figure 4: The confusion matrix between the true labels and the predicted labels from GPT 3.5-Relevant.

According to Table 3, we can see that both the Macro average precision and the Micro average precision of GPT 3.5 - Relevant are higher than those of GPT 3.5 - Random. This indicates that utilising relevant examples for demonstration leads to an improvement in overall performance. Examining the models' performance across individual categories, we observe that there is not a consistent improvement in all categories. In certain categories, such as "physical sciences (02)," "built environment and design (12)," "commerce, management, tourism, and services (15)," "studies in creative arts and writing (19)," and "language, communication, and culture (20)," incorporating relevant examples results in a significant improvement. Conversely, in some categories like "agricultural and veterinary sciences (07)," "education (13)," and "economics (14)," including relevant examples leads to a decline in performance.

In specific categories, such as "earth sciences (04)," "engineering (09)," "technology (10)," and "history and archaeology (21)," our models tend to exhibit very poor performance. To identify the underlying causes, we analyse problematic cases through a confusion matrix, as shown in Figure 4, and conduct human evaluations.

Referring to Figure 4, we can observe that GPT 3.5 models have low precision due to their difficulty in distinguishing these categories from several other categories. Taking "earth sciences (04)" as an example, GPT 3.5-Relevant predicts 96 "engineering (09)" records, 21 "mathematical sciences (01)" records, 10 "environmental sciences (05)" records, 10 "agricultural and veterinary sciences (07)" records as "earth sciences (04)." Upon close examination of the 96 "engineering (09)" records that were inaccurately predicted, we discovered that 94 out of the 96 had been labelled as "geodesy" by human annotators at the sub-group level (6-digit code). It's worth noting that labels at the sub-group level offer a higher degree of specificity. In the latest version of ANZSRC-FoR, ANZSRC-FoR 2020, "geodesy" falls under the sub-group "geophysics," and "geophysics" is one of groups of "earth sciences". So "geodesy" is moved from "engineering (09)" to "earth sciences (04)" in ANZSRC-FoR 2020. This means that our model's prediction ("earth sciences") is accurate if we choose the ANZSRC-FoR 2020 as the classification schema, while the prediction is wrong if we use the ANZSRC-FoR 2018. Additionally, we looked at the rest of the incorrectly predicted examples, we found that most of them are multidisciplinary or somewhat related to "earth sciences," such as examples given in Table 4.

Table 4: Example datasets from other categories that have been mistakenly classified as Earth Sciences (04).

| True Label | Dataset Title | Dataset Description |
|---|---|---|
| Mathematical Sciences (01) | Beagle Reef 1:50 000 topographic map | Product Specifications Coverage: Partial coverage, predominantly in northern Australia, along major transport routes, and other selected areas. About 1000 maps have been published to date. Currency: Ranges from 1968 to 2006. Coordinates: Geographical and UTM. Datum: AGD66, new edition WGS84; AHD. Projection: Universal Transverse Mercator UTM. Medium: Paper, flat copies only. |

| Environmental Sciences (05) | Oceanic Shoals Commonwealth Marine Reserve - High Resolution Multibeam Acoustic Backscatter Grids | This resource contains bathymetry and backscatter data for the Oceanic Shoals Commonwealth Marine Reserve (CMR) in the Timor Sea collected by Geoscience Australia during September and October 2012 on RV Solander (survey GA0339/SOL5650). The survey used a Kongsberg EM3002 300 kHz multibeam sonar system mounted in single head configuration to map four areas, covering a combined area of 507 square kilometres. Data are gridded to 2 m spatial resolution. The Oceanic Shoals Commonwealth Marine Reserve survey was undertaken as an activity within the Australian Government's National Environmental Research Program Marine Biodiversity Hub and was the key component of Research Theme 4 - Regional Biodiversity Discovery to Support Marine Bioregional Plans (The paragraph is overly lengthy, and as a result, the remainder has been excluded.) |
|---|---|---|
| Agricultural and Veterinary Sciences (07) | Stanley Fowler Image Collection | A selection of photographs from those taken by Stanley Fowler (1895-1961).<br /> With the co-operation of the Royal Australian Air Force, in 1936-39 and 1942-46 he undertook a series of aerial observations of pelagic stock in coastal waters. His photographs number over 10,000.<br /> This collection includes photographs of landforms, coastal areas, vessels, equipment and people from South Australia, Queensland, Victoria, Tasmania and Flinders Island. It also includes photographs of the research vessel, "Warreen", in which the C.S.I.R.'s decision to purchase Fowler had been instrumental. |

Since most of the "engineering (09)" records were predicted as "earth sciences (04)," this had a significant impact on the precision of the "engineering (09)" category. "Technology (10)" records tend to be more multidisciplinary, resulting in predictions spanning other categories like "chemical sciences (03)" and "biological sciences (06)." It has been noted that "technology (10)" has been removed from ANZSRC-FoR 2020. For both "engineering (09)" and "earth sciences (04)," our GPT 3.5 - Random model exhibits zero precision. The reasons are the same as those affecting GPT 3.5 - Relevant. However, GPT 3.5 - Random tends to predict all records from these two categories as belonging to other fields. This underscores the importance of including relevant examples as demonstration examples, as demonstrated by the improvement achieved by GPT 3.5 - Relevant in the results. The last category that our models perform very badly is "history and archaeology (21)," which happens to be the largest record group in RDA collections. Upon manual examination, we discovered that the "inaccurately" predicted labels from all test records encompassed 16 ANZSRC divisions. Our manual inspection revealed that the predominant scenario involved datasets categorised as "history and archaeology" because they are historical artefacts or archived records, yet their content displayed a connection to a particular research field.

For instance, the following is the metadata of a dataset from "history and archaeology":

> Title: '*Correspondence, Reports and Memoranda Regarding the Bulk Handling of Grain*'
> Description: '*This series is comprised of correspondence, reports and memoranda regarding the bulk handling of grain. Included in this series are reports of Royal Commissions and Premiers Conferences on the question of bulk handling of grain; letters and reports from private companies; comments of the Victorian Railways Commissioners; construction notes on proposed silos; and pamphlets and journals.*'
> 'Data_source_key': 'prov.vic.gov.au',
> 'date_from': ['1902-01-01T00:00:00.000Z'],
> 'date_to': ['1936-01-01T00:00:00.000Z']

The human label is "history and archaeology," while the predicted label is "agricultural and veterinary sciences" which is related to the content of the dataset. Note that our current method only uses the title and the description of a dataset as the features. There might be other informative metadata that affect human annotation and can be leveraged for effective categorization, such as 'Data_source_key,' 'date_from,' and 'date_to.'

**In comparison with supervised models:**

A previous study [2] applied four supervised machine learning classification models, e.g., multinomial logistic regression (MLR), multinomial naive bayes (MNB), K Nearest Neighbors (KNN), and Support Vector Machine (SVM), to the RDA collection. Because they are supervised machine learning models and using different testing data split, we refrain from directly comparing the performance of their models with ours. Nevertheless, we can draw a comparison between their best model (MLR) with our best model (GPT 3.5 - Relevant) regarding the categories encountering less or more difficulties with automatic classification methods. According to Table 5, there are categories in which both MLR and GPT 3.5 - Relevant excel, e.g., "physical sciences" and "studies in creative arts and writing," as well as categories in which both encounter challenges, e.g., "mathematical sciences (01)" and "technology."

Table 5: the comparison between MLR and GPT 3.5 - Relevant

| MLR is good at | GPT 3.5 - Relevant is good at | MLR is facing challenges in | GPT 3.5 - Relevant is facing challenges in |
|---|---|---|---|
| Physical sciences (02) Earth sciences (04) Biological sciences (06) Engineering (09) Studies in creative arts and writing (19) Language, communication and culture (20) History and archaeology (21) | Physical sciences (02) Chemical sciences (03) Built environment and design (12) Studies in creative arts and writing (19) Language, communication and culture (20) | Mathematical sciences (01) Information and computing sciences (08) Technology (10) Economics (14) Commerce, management, tourism and services (15) Psychology and cognitive sciences (17) | Mathematical sciences (01) Earth sciences (04) Engineering (09) Technology (10) History and archaeology (21) |

## 6. LIMITATIONS

Although our prompts with GPT-3.5 can achieve promising results, there are certain research fields where the model exhibits notably poor performance. One possible reason is that many research datasets are multidisciplinary, which poses challenges when ordering them according to their relevance. This also implies that investigating multi-label classification with in-context learning for subject metadata annotation is worthwhile. Another potential explanation is that humans utilise implicit or explicit classification guidelines, as discussed in the previous section. The example dataset in the discussion is categorised as "history and archaeology" by humans due to its archival nature, whereas our model predicts it as "agricultural and veterinary sciences" based on its content being linked to that field. In order to learn these classification rules or patterns, a machine learning model must undergo training using a specific quantity of instances and rich metadata features. As indicated in Table 5, we can see that supervised machine learning models excel in categorising datasets of "history and archaeology". In-context learning models face difficulties in this aspect due to the limited information in the context and the unsupervised nature. Consequently, learning classification rules or patterns from the existing datasets tend to be a weakness of in-context learning models.

At the time we were researching this problem, we lacked access to GPT-4. Given that GPT-4 has demonstrated superior performance compared to previous LLMs and most state-of-the-art systems across traditional NLP benchmarks [7], testing it in our annotation task will be a component of our future research.

# 7. CONCLUSION

In this study, our focus is on exploring the application of in-context learning based on the large language model GPT-3.5 for the automated annotation of subject metadata in dataset records. Through the use of customised prompts and relevant demonstration examples, our annotation techniques yielded promising results. Our findings are consistent with earlier studies that utilised language models as general automated annotators in domains other than subject metadata annotation. However, since the success of this approach relies on both the knowledge of the LLM and the contextual information provided in the prompt, grasping classification patterns across the entire data distribution proves to be challenging.

## AUTHOR CONTRIBUTION STATEMENT

# REFERENCES

[1] Wu, M., Psomopoulos, F., Khalsa, S. J., & de Waard, A.: Data Discovery Paradigms: User Requirements and Recommendations for Data Repositories. Data Science Journal, 18(1) (2019)

[2] Wu, M., Liu, Y. H., Brownlee, R., & Zhang, X.: Evaluating Utility and Automatic Classification of Subject Metadata from Research Data Australia (2021)

[3] Suominen, O., Inkinen, J., & Lehtinen, M.: Annif and Finto AI: Developing and Implementing Automated Subject Indexing. JLIS. it 13.1: 265-282 (2022)

[4] Mikolov, T., Chen, K., Corrado, G., & Dean, J.: Efficient Estimation of Word Representations in Vector Space. In Proceedings of the International Conference on Learning Representations (ICLR'13) (2013)

[5] Kenton, J. D. M. W. C., & Toutanova, L. K.: BERT: Pre-training of Deep Bidirectional Transformers for Language Understanding. In Proceedings of NAACL-HLT (pp. 4171-4186) (2019)

[6] Brown, T., Mann, B., Ryder, N., Subbiah, M., et al.: Language Models are Few-shot learners. Advances in Neural Information Processing Systems, 33, 1877-1901 (2020)

[7] OpenAI. Gpt-4 technical report. https://cdn.openai.com/papers/gpt-4.pdf., (2023).

[8] Wang, S., Liu, Y., Xu, Y., Zhu, C., & Zeng, M.: Want To Reduce Labelling Cost? GPT-3 Can Help. In Findings of the Association for Computational Linguistics: EMNLP 2021 (pp. 4195-4205) (2021)

[9] Gilardi, F., Alizadeh, M., & Kubli, M.: Chatgpt Outperforms Crowd-workers for Text-annotation Tasks. arXiv preprint arXiv:2303.15056 (2023)

[10] Thapa, S., Naseem, U., & Nasim, M.: From Humans to Machines: can ChatGPT-like LLMs Effectively Replace Human Annotators in NLP tasks. In Workshop Proceedings of the 17th International AAAI Conference on Web and Social Media (2023)

[11] Liu, J., Shen, D., Zhang, Y., Dolan, W. B., Carin, L., & Chen, W.: What Makes Good In-Context Examples for GPT-3?. In Proceedings of Deep Learning Inside Out (DeeLIO 2022): The 3rd Workshop on Knowledge Extraction and Integration for Deep Learning Architectures (pp. 100-114) (2022)

[12] Wu, M., Brandhorst, H., Marinescu, M. C., Lopez, J. M., Hlava, M., & Busch, J.: Automated Metadata Annotation: What is and is not Possible with Machine Learning. Data Intelligence, 5(1), 122-138 (2023).

[13] Boyack, K. W., & Klavans, R.: Co-citation Analysis, Bibliographic Coupling, and Direct Citation: Which Citation Approach Represents the Research Front most Accurately?. Journal of the American Society for Information Science and Technology, 61(12), 2389-2404 (2010)

[14] Janssens, F., Zhang, L., De Moor, B., & Glänzel, W.: Hybrid Clustering for Validation and Improvement of Subject-classification Schemes. Information Processing & Management, 45(6), 683-702 (2009)

[15] Bubeck, S., Chandrasekaran, V., Eldan, R., Gehrke, J., et al.: Sparks of Artificial General Intelligence: Early Experiments with GPT-4. arXiv preprint arXiv:2303.12712 (2023)

[16] Chiang, C. H., & Lee, H. Y.: Can Large Language Models Be an Alternative to Human Evaluations?. arXiv preprint arXiv:2305.01937 (2023)

[17] Wei, J., Bosma, M., Zhao, V., Guu, K., Yu, A. W., Lester, B., ... & Le, Q. V.: Fine Tuned Language Models are Zero-Shot Learners. In the International Conference on Learning Representations (2021)

[18] Ouyang, L., Wu, J., Jiang, X., Almeida, D., Wainwright, C., Mishkin, P., ... & Lowe, R.: Training Language Models to Follow Instructions with Human Feedback. Advances in Neural Information Processing Systems, 35, 27730-27744 (2022)

[19] Honovich, O., Scialom, T., Levy, O., & Schick, T.: Unnatural Instructions: Tuning Language Models with (almost) no Human Labour. arXiv preprint arXiv:2212.09689 (2022)

[20] Veselovsky, V., Ribeiro, M. H., & West, R.: Artificial Artificial Artificial Intelligence: Crowd Workers Widely Use Large Language Models for Text Production Tasks. arXiv preprint arXiv:2306.07899 (2023)

[21] Liu, P., Yuan, W., Fu, J., Jiang, Z., Hayashi, H., & Neubig, G.: Pre-train, Prompt, and Predict: A Systematic Survey of Prompting Methods in Natural Language Processing. ACM Computing Surveys, 55(9), 1-35 (2023)

[22] Jiang, Z., Xu, F. F., Araki, J., & Neubig, G.: How Can We Know What Language Models Know?. Transactions of the Association for Computational Linguistics, 8, 423-438 (2020)

[23] Li, X. L., & Liang, P.: Prefix-Tuning: Optimizing Continuous Prompts for Generation. In Proceedings of the 59th Annual Meeting of the Association for Computational Linguistics and the 11th International Joint Conference on Natural Language Processing (Volume 1: Long Papers) (pp. 4582-4597) (2021)

[24] Wei, J., Wang, X., Schuurmans, D., Bosma, M., Xia, F., Chi, E., ... & Zhou, D.: Chain-of-thought Prompting Elicits Reasoning in Large Language Models. Advances in Neural Information Processing Systems, 35, 24824-24837 (2022)